\documentclass[10pt,twocolumn,letterpaper]{article}

\usepackage{cvpr}
\usepackage{times}
\usepackage{epsfig}
\usepackage{graphicx}
\usepackage{amsmath}
\usepackage{amssymb}
\usepackage{soul}
\usepackage{color}

\usepackage{booktabs}
\usepackage{siunitx}

\usepackage[breaklinks=true,bookmarks=false]{hyperref}

\cvprfinalcopy 

\def\cvprPaperID{****} 

\DeclareMathOperator*{\argmin}{argmin}

\usepackage[page]{appendix}
\usepackage{listings}

\definecolor{codegreen}{rgb}{0,0.6,0}
\definecolor{codegray}{rgb}{0.5,0.5,0.5}
\definecolor{codepurple}{rgb}{0.58,0,0.82}
\definecolor{backcolour}{rgb}{0.95,0.95,0.92}
 
\lstdefinestyle{mystyle}{
    backgroundcolor=\color{backcolour},   
    commentstyle=\color{codegreen},
    keywordstyle=\color{magenta},
    numberstyle=\tiny\color{codegray},
    stringstyle=\color{codepurple},
    basicstyle=\footnotesize,
    breakatwhitespace=false,         
    breaklines=true,                 
    captionpos=b,                    
    keepspaces=true,                 
    numbers=none,                    
    numbersep=5pt,                  
    showspaces=false,                
    showstringspaces=false,
    showtabs=false,                  
    tabsize=2
}
 
\begin{document}

\title{Visual Feature Attribution using Wasserstein GANs}




\author{Christian F. Baumgartner$^1$ 
\and Lisa M. Koch$^2$
\and Kerem Can Tezcan$^1$
\and Jia Xi Ang$^1$ 
\and Ender Konukoglu$^1$
\and for the Alzheimer's Disease Neuroimaging Initiative\thanks{Data used in preparation of this article were obtained from the Alzheimer’s Disease
Neuroimaging Initiative (ADNI) database (adni.loni.usc.edu).}%
\vspace{1pt}
\and $^1$Computer Vision Lab, ETH Zurich \\
\and $^2$Computer Vision and Geometry Group, ETH Zurich} 

\maketitle

\begin{abstract}
Attributing the pixels of an input image to a certain category is an important and well-studied problem in computer vision, with applications ranging from weakly supervised localisation to understanding hidden effects in the data. In recent years, approaches based on interpreting a previously trained neural network classifier have become the de facto state-of-the-art and are commonly used on medical as well as natural image datasets. In this paper, we discuss a limitation of these approaches which may lead to only a subset of the category specific features being detected. To address this problem we develop a novel feature attribution technique based on Wasserstein Generative Adversarial Networks (WGAN), which does not suffer from this limitation. We show that our proposed method performs substantially better than the state-of-the-art for visual attribution on a synthetic dataset and on real 3D neuroimaging data from patients with mild cognitive impairment (MCI) and Alzheimer's disease (AD). For AD patients the method produces compellingly realistic disease effect maps which are very close to the observed effects. 
\end{abstract}

\section{Introduction} 

In this paper we address the problem of visual attribution, which we define as detecting and visualising evidence of a particular category in an image. Pinpointing all evidence of a class is important for a variety of tasks such as weakly supervised localisation or segmentation of structures \cite{oquab2015object,pinheiro2015image,zhou2016learning}, and better understanding disease effects, and physiological or pathological processes in medical images \cite{zhu2017deep,ge2017skin,feng2017discriminative,gondal2017weakly,fong2017interpretable,jamaludin2016spinenet,sundararajan2017axiomatic,konukoglu2016constructing,konukoglu2017subcmap,zhang2017weakly}. 

Currently, the most frequently used approach to address the visual attribution problem is training a neural network classifier to predict the categories of a set of images and then following one of two strategies: analysing the gradients of the prediction with respect to an input image \cite{jamaludin2016spinenet,baumgartner2017sononet,sundararajan2017axiomatic} or analysing the activations of the feature maps for the image \cite{zhou2016learning,oquab2015object,pinheiro2015image} to determine which part of the image was responsible for making the associated prediction. 


\begin{figure}[t]
  \centering
    \includegraphics[width=\columnwidth]{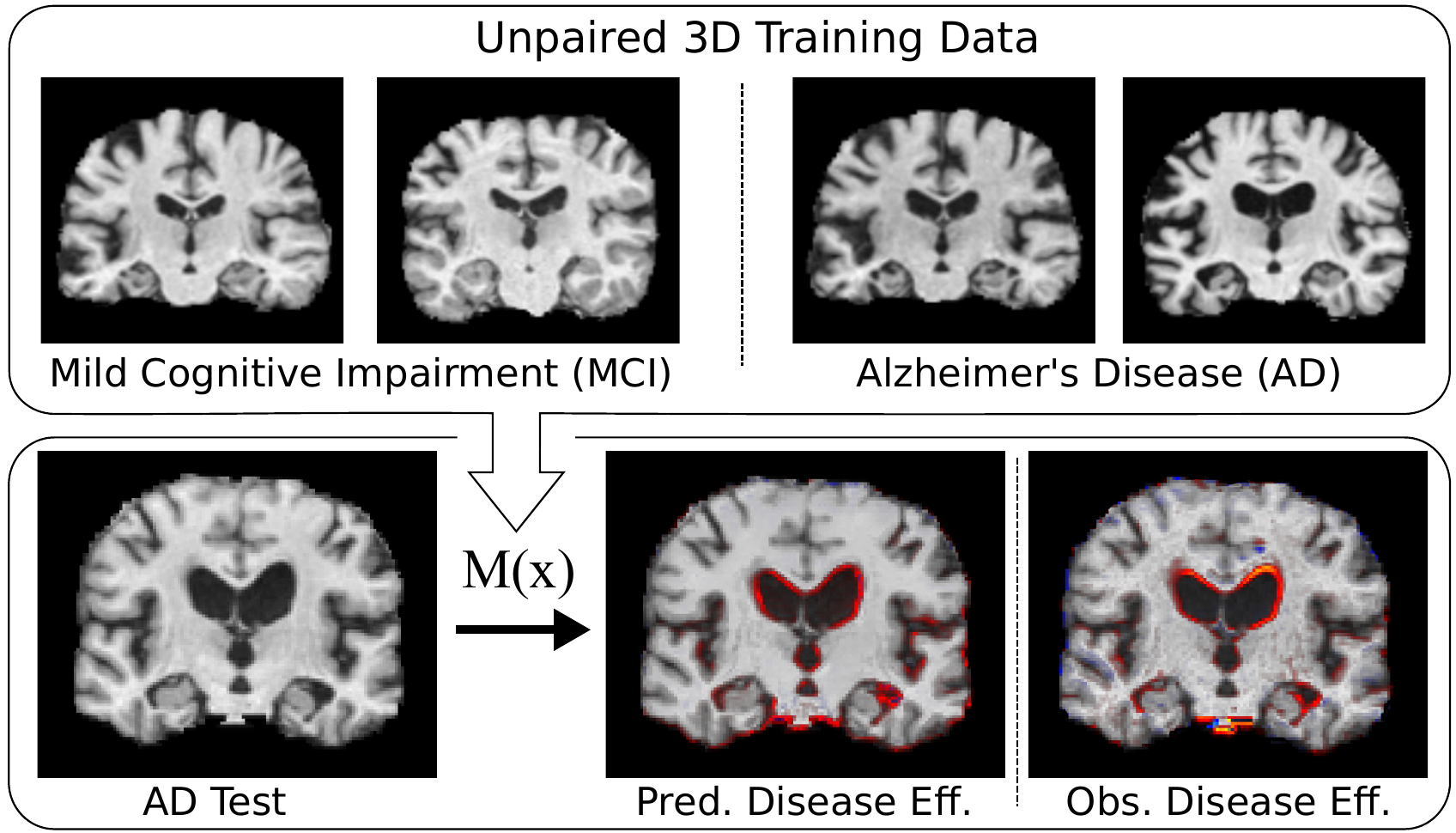}
    \caption{Our proposed method learns a map generating function $M(x)$ from unlabelled training data. Given a test image, this function will generate an image-specific visual attribution map which highlights the features unique to that category. The method is of particular interest for creating medical disease effect maps. We show that on neuroimaging data the method predicts effects in very good agreement with the actual observed effects.}
    \label{fig:what}
\end{figure}

Visual attribution based directly on neural network classifiers may, under some circumstances, produce undesired results. It is known that such classifiers base their decisions on certain salient regions rather than the whole object of interest. It was recently shown that during training neural networks minimise the mutual information between input and output layers, thereby compressing the input features \cite{shwartz2017opening}. These findings suggest that a classifier may ignore features with low discriminative power if stronger features with redundant information about the target are available. In other words, neural network training may be working in opposition to the goal of visual attribution. As a consequence, if there is evidence for a class at multiple locations in the image (such as multiple lesions in medical images) some locations may not influence the classification result and may thus not be detected. We demonstrate this effect on a synthetic dataset in our experiments. 

It would be highly desirable if instead we could visualise evidence of a particular category in a way that captures \emph{all} category-specific effects in an image. Our main contribution is a novel approach towards solving the visual attribution problem which takes a first step in this direction. In contrast to the majority of recent techniques, the method does not rely on a classifier but rather aims at finding a map that, when added to an input image of one category, will make it indistinguishable from images from a baseline category. To this end we propose a generative model in which the additive map is learned as a function of the images. The method is based on Wasserstein generative adversarial networks (WGAN) \cite{arjovsky2017wasserstein}, which have the desirable property that they minimise an approximation of the Wasserstein distance between the distributions of the generated images and the real ones. 

We note that our method does not tackle the classification problem but rather assumes that the category labels of the test images have already been determined (e.g. using a separately trained classifier or by an expert). Furthermore, the method requires a baseline category, which is not the case for many benchmark recognition datasets in vision, but is in fact the case for many practical detection applications, especially in medical image analysis. 

We demonstrate the method on synthetic 2D data and on large 3D brain MR data, where we aim to predict subject-specific disease effect maps for Alzheimer's disease (AD).

\subsection{Medical motivation}

 Identifying disease effects at the subject-specific level is of great interest for various medical applications. In clinically oriented research, identifying subject-specific disease effects would be useful for stratification amongst the patient population and to help disentangling diseases such as AD~\cite{iqbal2005subgroups} and Schizophrenia~\cite{ross2006neurobiology}, that are believed to be composed of multiple sub-types rather than a single disease. Furthermore, for clinicians, subject-specific maps could be helpful in assessing disease status and grading. 

In this paper, we chose to study the disease effects of AD with respect to mild cognitive impairment (MCI), which is characterised by a slight decline in cognitive abilities. Patients with MCI are at increased risk of developing AD, but do not always do. We evaluate our method on one of the largest publicly available neuroimaging datasets acquired by the Alzheimer's Disease Neuroimaging Initiative (ADNI). We used the MCI population as the baseline category and the AD population as the category of interest. Our choice to use MCI as our baseline is motivated by the fact that the ADNI dataset contains a number of MCI subjects who convert AD with imaging data at both stages of the disease. This allowed us to evaluate the predicted disease effects against real observed effects defined as the differences between images at the different stages. Note that even though using normal controls as the baseline is feasible, it would have been much harder to assess the proposed method due to the small number of control to AD converters in the ADNI dataset.



\section{Related work}

\subsection{Visual attribution}

A commonly used approach for weakly supervised localisation or segmentation is to analyse the final feature map of a neural network classifier \cite{oquab2015object,pinheiro2015image}. The Class Activation Mapping (CAM) method \cite{zhou2016learning} builds on those techniques by reducing the feature maps of the second to last layer using a global average pooling layer, followed by a dense prediction layer. This allows to create class-specific activation maps as a linear combination of the weights in the last layer.

A large amount of works on medical images builds on the CAM technique. Examples include the work of Feng et al. \cite{feng2017discriminative} on pulmonary nodule localisation in CT, the work of Ge et al. \cite{ge2017skin} on skin disease recognition. Other examples are \cite{zhu2017deep}, \cite{gondal2017weakly}. It is important to note that CAM is restricted in the resolution of its visual attributions by the resolution of the last feature map. Consequently, often post-processing of the predictions is required \cite{feng2017discriminative,fong2017interpretable,pinheiro2015image}. In contrast, our proposed method can produce visual attributions at the resolution of the original input images.

Another class of techniques creates saliency maps by backpropagating back to the input image. Examples include Guided Backprop \cite{springenberg2014striving}, Excitation Backprop \cite{zhang2016top}, Integrated Gradients \cite{sundararajan2017axiomatic}, meaningful perturbations \cite{fong2017interpretable}. 

Similar techniques have been applied in the domain of medical images. Jamaludin et al. \cite{jamaludin2016spinenet} use the backprop-based saliency technique proposed by \cite{simonyan2013deep} to pinpoint lumbar degradations, and Baumgartner et al. \cite{baumgartner2017sononet,baumgartner2016real} use a variant of \cite{springenberg2014striving} to localise fetal anatomy. Gao and Noble \cite{gao2017detection} apply a similar approach to localise the fetal heart.


\subsection{Statistical disease models}
Statistical analysis of medical images for identifying disease effects has been an instrumental tool for various diseases and disorders~\cite{thompson2001cortical, rosas2002regional, burton2004cerebral} as well as other non-disease related factors~\cite{garrido1993cortical, miller2003effects, kanai2011structural, watkins2002mri, peper2007genetic, thompson2001genetic}. The most common approach is to use regression analysis or machine learning tools to generate population average maps, which highlights features that are salient across the population~\cite{ashburner2001voxel,krishnan2011partial,worsley1997characterizing,gaonkar2013analytic,mwangi2014review, rahim2015integrating,ganz2015relevant}.

Recently, constructing subject-specific maps has received attention. Maumet et al. took a one-versus-all group analysis approach~\cite{Maumet:2013ha, Maumet:2016kv}, while Konukoglu and Glocker extracted subject-specific maps with predictive models and Markov Random Field restoration~\cite{konukoglu2016constructing, konukoglu2017subcmap}.

The common drawback in the previous approaches is the need for \emph{registration}. In order to compute disease effect maps, images of different subjects need to be non-rigidly aligned on a common template where statistical analysis can be performed. The non-rigid registration process brings additional uncertainty to the subject-specific maps. Our work, addresses this shortcoming and generates subject-specific disease effect maps without requiring registration. 

\subsection{Image generation using GANs}

Generative adversarial images conditioned on an input image have been used in diverse applications such as video frame prediction \cite{mathieu2015deep}, image super-resolution \cite{ledig2016photo}, image-translation across domains using paired \cite{isola2016image} and unpaired \cite{zhu2017unpaired} images, and pixel level domain adaptation \cite{bousmalis2016unsupervised,shrivastava2016learning}.

In the context of medical images, GANs have been applied to super-resolution in retinal fundus images \cite{mahapatra2017image}, for semi-supervised cardiac segmentation \cite{zhang2017deep}, synthesising computed tomography images from MR images\cite{nie2017medical,wolterink2017deep} and intraoperative motion modelling \cite{hu2017intraoperative}. Although some of the above models use 3D data, the examined volumes are usually relatively small \cite{hu2017intraoperative}, or the networks operate in a patch-wise fashion \cite{nie2017medical}.  It is important to note that in the case of brain MR images of Alzheimer’s disease patients, the diagnostic information is only visible at a high resolution and cannot be determined by considering small local patches only. In this paper, we therefore tackle the challenge of processing large 3D volumes directly.

\subsection{Contributions}

\begin{enumerate}
  \item We demonstrate a limitation in current neural network based visual attribution methods using synthetic data. 
  \item We propose a novel visual attribution technique that can detect class specific regions more completely and at a high resolution. 
  \item To our knowledge, this is the first application of generative adversarial networks on large structural 3D data. 
\end{enumerate}

An implementation of the proposed method is publicly available here: \url{https://github.com/baumgach/vagan-code}.

\section{Visual attribution using WGANs}

\subsection{Problem Formulation}

\begin{figure*}[t]
  \centering
    \includegraphics[width=0.95\textwidth]{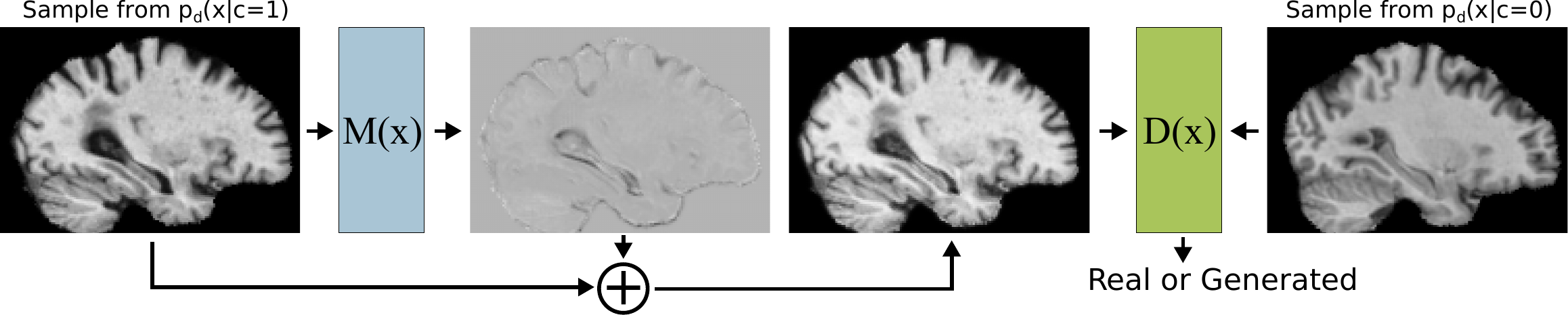}
    \caption{Overview of VA-GAN. During training images are sampled from the categories $c\in\{0,1\}$. Images from $c=1$ are passed to the map generating function $M(x)$. The map generator aims to create additive maps which produce generated images that the critic $D(x)$ cannot distinguish from images sampled from $p_d(x|c=0)$. The critic, $D(x)$ tries to assign different values to generated and real images. During testing, $M(x)$ can be used directly to predict a map in a single forward pass.}
    \label{fig:method}
\end{figure*}

Our goal is to estimate a map that highlights the areas in an image which are specific to the class the image belongs to. We formulate the problem for two classes $c \in \{0,1\}$, a baseline class and a class of interest. The formulation however, easily extends to the case of multiple classes of interest. We denote an image with $x$ and the distribution of images coming from class $c=0$ with $p_{d}(x|c=0)$ and images from class $c=1$ with $p_{d}(x|c=1)$. In the case of medical application, $c=1$ could for example denote the set of images from a population with a certain disease and $c=0$ images of control subjects.

We formulate a problem as estimating a map function $M(x)$ that, when added to an image $x_i$ from category ${c=1}$, creates an image 
\begin{equation}\label{eq:problem}
y_i = x_i + M(x_i),
\end{equation} 
which is indistinguishable from the images sampled from ${p_{d}(x|c=0)}$. Thereby, the map $M(x_i)$ contains all the features which distinguish the input image $x_i$ from the other category. In the case of medical images, $M$ will by definition contain the effects of a disease visible in the images, i.e. a \emph{disease effect map}. 

We model the function $M$ using a convolutional neural network, whose parameters we find using a WGAN. 

\subsection{Wasserstein GANs}

In the GAN paradigm a generator function and a discriminator function (both neural networks) compete with each other in a zero-sum game \cite{goodfellow2014generative}. Given random noise as input, the generator tries to produce realistic images that fool the discriminator, while the discriminator tries to learn the difference between generated and real images. 

Arjovski and Bottou pointed out a limitation in this paradigm which precludes a guarantee that the generated images will necessarily converge to the target distribution \cite{arjovsky2017towards} (although in practice, with appropriate training methods, many impressive results were achieved \cite{radford2015unsupervised}). Wasserstein GANs are a modification to the classic GAN paradigm where the discriminator is replaced by a critic which does not have an activation function in its final layer and which is constrained to be a $K$-Lipschitz function. WGANs have better optimisation properties and it can be shown that they minimise a meaningful distance between the generated and real distributions. 

\subsection{Constrained effect maps using WGANs}

In this work we build on WGANs to find the optimal map generation function. In contrast to regular WGANs, we have a map generator function $M(x_i)$, which, during training, takes as input randomly sampled images $x_i$ from category $c=1$ rather than noise. $M$ tries to generate maps that, when added to $x_i$, create images $y_i$ appearing to be from category $c=0$. By trying to distinguish generated images $y_i$ from real images from category $c=0$, the critic $D$ ensures that the generated maps are constrained to realistic modifications (see Fig. \ref{fig:method} for an overview). In the context of medical images, this means enforcing anatomically realistic modifications to the images. 

Building on \cite{arjovsky2017wasserstein} this leads to the following cost function:
\begin{equation}\label{eq:wgan}
\begin{split}
\mathcal{L}_{GAN}(M,D) & = \mathbb{E}_{x\sim p_d(x|c=0)}[D(x)] \\ & - \mathbb{E}_{x \sim p_d(x|c=1)}[D(x + M(x))].
\end{split}
\end{equation}

Optimising Eq. \ref{eq:wgan} directly could lead to changes in the input image $x_i$ that change the image identity. For instance, the brain anatomy of a subject could be changed to a degree where it does not only capture disease related changes but changes the subject identity. We want to encourage the smallest required map $M$ that still leads to a realistic $y_i$. Thus add the following data regularisation term to the cost function:
\begin{equation}
  \mathcal{L}_{reg}(M) = ||M(x)||_1,
\end{equation}
where $||\cdot||_1$ is the L1 norm \cite{fong2017interpretable}.

The final optimisation is then given by
\begin{equation}\label{eq:total_cost}
M^* = \argmin\limits_{M} \max\limits_{D\in\mathcal{D}} \mathcal{L}_{GAN}(M,D) + \lambda \mathcal{L}_{reg}(M),
\end{equation}
where $\mathcal{D}$ is the set of 1-Lipschitz functions. 

In order to enforce the Lipschitz constraint we use the optimisation method proposed in \cite{gulrajani2017improved}. As recommended by \cite{gulrajani2017improved}, we weigh the gradient penalty with a factor of 10 throughout all experiments. 
 
\subsection{Network architecture}

As we will discuss in more detail in Section \ref{sec:exp_real}, we design our proposed method with large 3D medical imaging data in mind, which often need to be processed at high resolutions in order to retain diagnostic information. Specifically, in our experiments on neuroimaging data, an input volume size of 128x160x112 voxels is used. 

With such large images the limiting factor becomes storing the activations of the networks on GPU memory. With this in mind we design the map generator and the critic networks as follows. 

\subsubsection{Map generator network}

The map generator function should be able to form an internal representation of the visual attributes that characterise the categories. In the case of brain images affected by dementia, it should be able to ``understand'' the systematic changes involved in the disease. Therefore, a relatively powerful network is required to adequately model the function $M$. To this end, we use the 3D U-Net \cite{cciccek20163d} (originally proposed for segmentation), as a starting point. The 3D U-Net has an encoder-decoder structure with a bottle-neck layer in the middle, but additionally introduces skip connections at each resolution level bypassing the bottle-neck. This allows the network to combine high-level semantic information (such as the presence of a structure) with low-level information (such as edges). 


In order to reduce GPU memory consumption we reduce the number of feature maps by a factor of 4 in most layers. As in the original 3D U-Net \cite{cciccek20163d} we use batch normalisation for all layers except the final one. The exact architecture is shown in Fig. 2 in the supplementary material.

\subsubsection{Critic function} 

In line with related literature on image generation using GANs \cite{isola2016image,zhu2017unpaired,shrivastava2016learning}, we model our critic as a fully convolution network with no dense layers. We loosely base our architecture on the C3D network which achieved impressive results on action recognition tasks in video data by processing them directly in the spatio-temporal 3D space \cite{tran2015learning}. However, in contrast to that work we only perform 4 pooling steps. After the fourth pooling layer we add another 3x3x3 convolution layer, followed by a 1x1x1 convolution layer which reduces the number of feature maps to one. The final critic prediction is given by a global average pooling operation of that feature map. 

It proved important \emph{not} to use batch normalisation for the critic network. Towards the beginning of training generating statistics of a batch with generated and the real images may not produce reasonable estimates, because the images vary considerably from each other. We surmise that this effect prevents the critic from learning when batch normalisation is used. A similar observation was made in \cite{gulrajani2017improved}. We also experimented with layer normalisation \cite{ba2016layer}, but did not observe improvements. 

The exact architecture we used is shown in Fig. 1 in the supplementary material.

\subsection{Training}

To optimise our networks, we follow \cite{arjovsky2017wasserstein,gulrajani2017improved} and update the parameters of the critic and map generator networks in an alternating fashion. In contrast to the regular GANs \cite{goodfellow2014generative}, WGANs require a critic which is kept close to optimality through-out training. We therefore perform 5 critic updates for every map generator update. Additionally, for the first 25 iterations and every hundredth iteration, we perform 100 critic updates per generator update. 

With the above architectures, the maximum batch size that can be used for a single gradient computation on a Nvidia Titan Xp GPU with 12 GB of memory is 2+2 (real+generated). In order to obtain more reliable gradient estimates we aggregate the gradients for a total of 6 mini-batches before performing a training step. 

We used the ADAM optimiser \cite{kingma2014adam} to perform the update steps for all experiments. The optimiser parameters were set to $\beta_1=0$, $\beta_2=0.9$, and we used a learning rate of $10^{-3}$. Lastly, we used a weight of $\lambda=10^2$ for the map regularisation term (see Eq. \ref{eq:total_cost}) throughout the paper. Training took approximately 24 hours on an Nvidia Titan Xp.  

\section{Experiments}

We evaluated the proposed method using a synthetically generated dataset and a large number of 3D brain MRI images from the publicly available ADNI dataset. 




We compared our proposed visual attribution GAN (VA-GAN) to methods from the literature which have been used for visual attribution both on natural and on medical images. Specifically, we compared against Guided Backpropagation \cite{springenberg2014striving}, Integrated Gradients \cite{sundararajan2017axiomatic} and Class Activation Mapping (CAM) \cite{zhou2016learning}. Furthermore, to verify that the WGAN framework is necessary, we also investigated an alternative way of estimating the additive map not based on GANs, which is described in detail in the next section. 

All the methods except VA-GAN use classification networks. For simplicity, we used a very similar architecture for these networks as for the critic in VA-GAN, except for two differences: (1) we replaced the last convolution and the global average pooling layer by two dense layers followed by a softmax and (2) we used batch normalisation for all layers, which produced better classification results for the experiments on the ADNI dataset. In addition, for the CAM method we designed the last layer as described by \cite{zhou2016learning} and omitted the last two max pooling layers, which allowed significantly more accurate visual attribution maps due to the higher resolution of the last feature maps. 

Lastly, for the experiments on the 2D synthetic data we simply replaced all 3D operations by 2D operations, but left the architectures otherwise unchanged. 

\subsection{Classifier-based map estimation}

In the VA-GAN approach, we generate an additive map which is constrained by the critic to generate a realistic image from the opposite class. To demonstrate that this approach is necessary we also investigated an alternative method of estimating the additive map without a term enforcing realistic maps. 

The alternative approach requires training a classifier $p(c=1) = f(\cdot)$ and then optimising an additive map \emph{image} $m$ that lowers the prediction $p(c=1)$ as much as possible. That is to say, the image $y_i = x_i + m$ should minimise $f_i(y_i)$. This formulation is almost exactly the same as for the WGAN-based approach (see Eq. \ref{eq:problem}) except that $m$ is not a function of $x_i$. 

We need to use a regularisation in determining $m$ to avoid trivial solutions, such as imperceptible changes that can fool classifiers~\cite{goodfellow2014explaining}. A ``well behaved'' map can be found by the following minimisation problem: 
\begin{equation}
\begin{split}
\label{eq:constructive_perturbation}
m^* = \argmin\limits_{m} f(x_i + m) & + \omega_1 ||m||_1 \\ & + \omega_2 \sum_u||\nabla m(u)||^\beta_\beta. 
\end{split}
\end{equation}
Here $u$ indexes the pixels or voxels of $m$. The L1 term weighted by $\omega_1$ encourages small maps, while the total variation term weighted by $\omega_2$ encourages smoothness. 

We optimise this cost function using the ADAM optimiser using the default internal parameters given in \cite{kingma2014adam} with a learning rate of $10^{-2}$ and early stopping at 1500 iterations. Furthermore, we set $\beta=2.0$, $\omega_1=10^{-2}$ and $\omega_2=10^{-5}$ in all experiments. 

This approach is strongly related to the meaningful perturbation masks technique proposed by \cite{fong2017interpretable} in which parts of an image are locally deleted by a mask $m$ such that the prediction $f(x)$  is minimised. In preliminary experiments we found that on the medical image problem we studied, visual attribution using destructive masks did not lead to the desired results. Deleting the diagnostic part of an image will not produce an image of the opposite class but rather an image with an undetermined diagnosis. This means such a mask may contain information about the location of diagnostic regions but not about specific disease effects, e.g. enlargement or shrinkage. In contrast, by optimising Eq. \ref{eq:constructive_perturbation} we attempt to morph the image into the opposite class, such that diagnostic regions can be changed to have the characteristics of another class. Because of the similarity to \cite{fong2017interpretable}, we refer to this method as additive perturbation maps.  

\subsection{Synthetic experiments}


\noindent{\bf Data: } In order to quantitatively evaluate the performance of the examined visual attribution methods, we generated a synthetic dataset of 10000 112x112 images with two classes, which model a healthy control group (label 0) and a patient group (label 1). The images were split evenly across the two categories. We closely followed the synthetic data generation process described in \cite{konukoglu2017subcmap} where disease effects were studied in smaller cohorts of registered images.  

\begin{figure}[t]
  \centering
    \includegraphics[width=0.78\columnwidth]{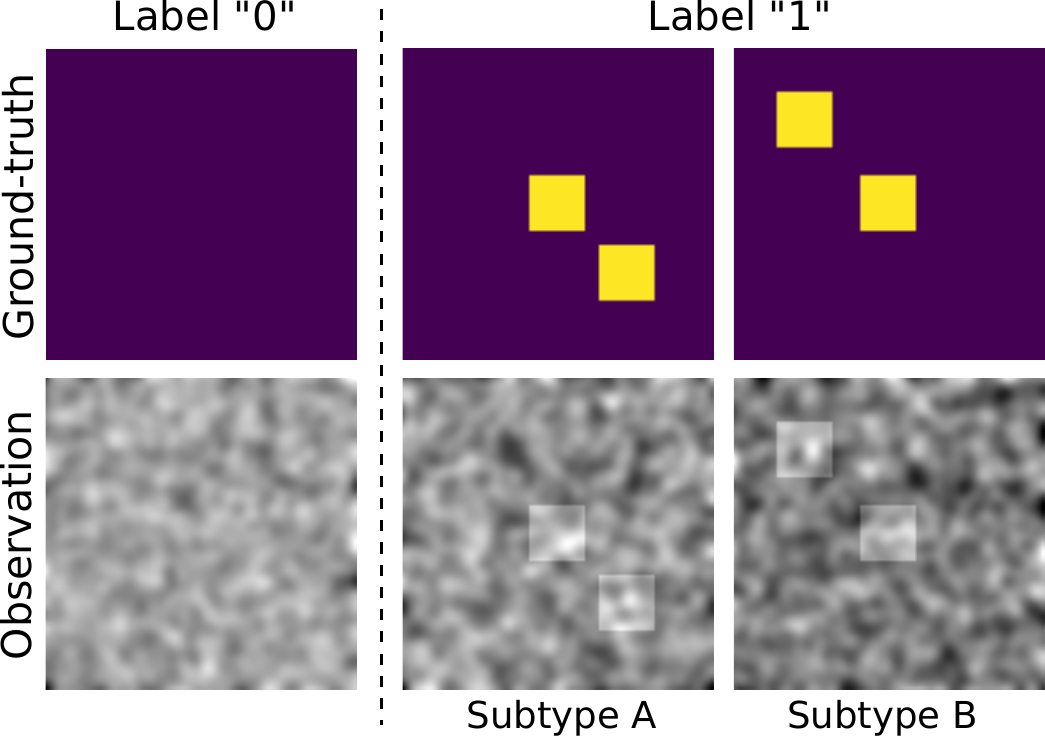}  
    \caption{Description of synthetic data. We generated noisy observations from ground-truth effect maps. The dataset contained two categories: A baseline category 0 (e.g. healthy images) and category with an effect (e.g. patient images). The images in category 1 contained one of two subtypes, A or B, which is unknown to the algorithms. A: box in the lower right, B: box in the upper left.}
    \label{fig:synth_data}
\end{figure}

The control group (label 0) contained images with random iid Gaussian noise convolved with a Gaussian blurring filter. Examples are shown in Fig. \ref{fig:synth_data}. The patient images (label 1) also contained the noise, but additionally exhibited one of two disease effects which was generated from a ground-truth effect map: a square in the centre and a square in the lower right (subtype A), or a square in the centre and a square in the upper left (subtype B). Importantly, both disease subtypes shared the same label. The location of the off-centre squares was randomly offset in each direction by a maximum of 5 pixels. This effect was added to make the problem harder, but had no notable effect on the outcome. \\ \vspace{-3mm}



\noindent{\bf Evaluation: } We split the data into a 80-20 training and testing set. Moreover, we used 20\% of the training set for monitoring the training. Next, we estimated the disease effect maps for all cases from the synthetic patient class using the examined methods. 

In order to assess the visual attribution accuracy quantitatively, we calculated the normalised cross correlation (NCC) between the ground-truth label maps and the predicted disease effect maps. The NCC has the advantage that it is not sensitive to the magnitude of the signals. For CAM we used only the positive values to calculate the NCC, while for the backprop-based techniques we used the absolute value, since those techniques do not necessarily predict the correct sign of the changes.\\ \vspace{-3mm}

\begin{figure}[t]
  \centering
    \includegraphics[width=\columnwidth]{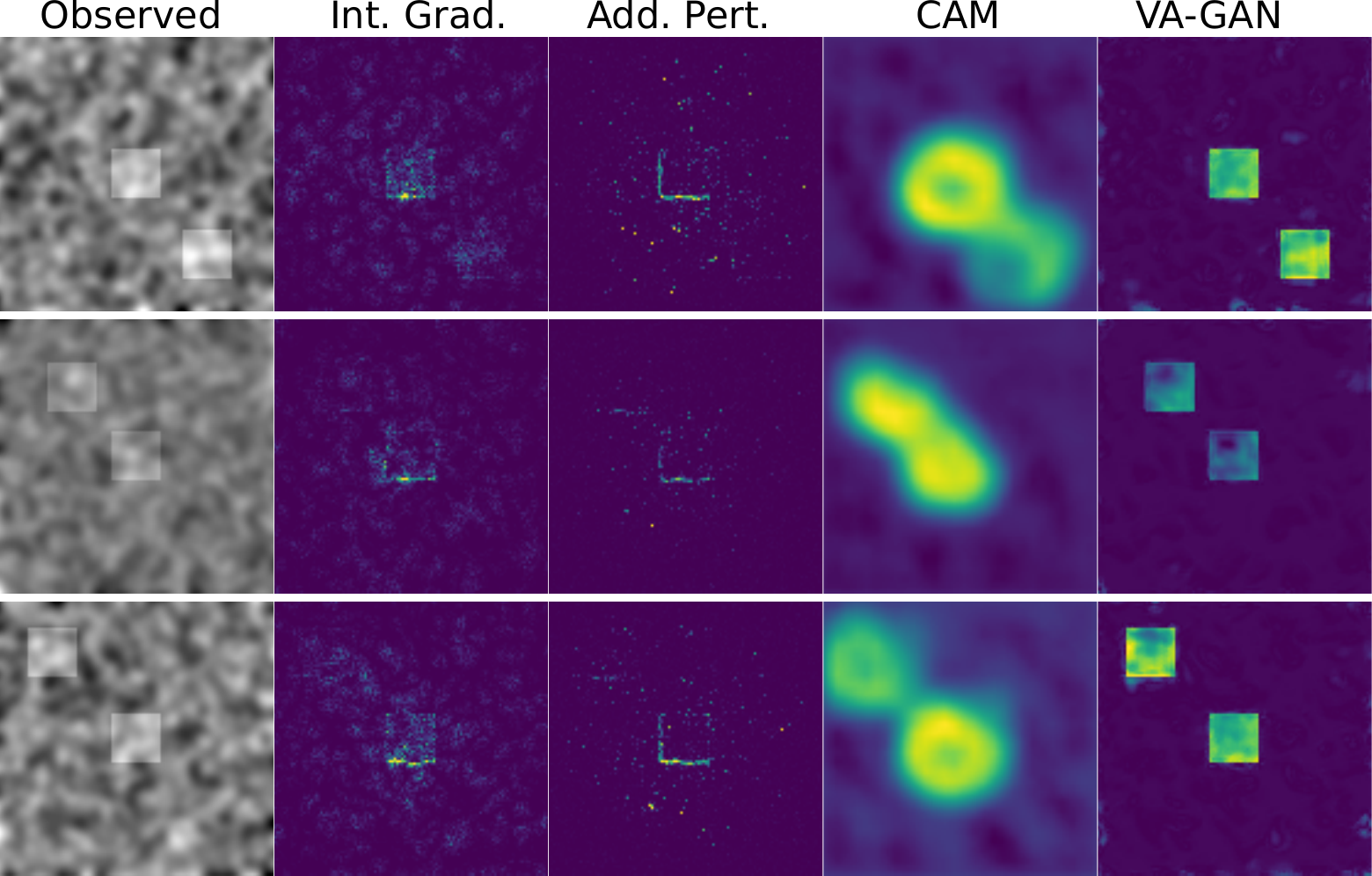}
    \caption{Examples of visual attribution on synthetic data obtained using the compared methods.}
    \label{fig:synth_results}
\end{figure}

\noindent{\bf Results: } A number of examples of the estimated disease effect maps are shown in Fig. \ref{fig:synth_results}. Guided Backpropagation produced similar results to Integrated Gradients. We therefore omitted it in the visual results due to space considerations but provide quantitative results. 

For the backprop-based methods we consistently observed two behaviours: 1) They tended to focus exclusively on the central square which was always present and was thus the most predictive set of features. This behaviour is consistent with the network compressing away less predictive features discussed earlier \cite{shwartz2017opening}. 2) They tended to focus mostly on the edges of the boxes rather than on the whole object. This may have to do with the fact that edges are more salient than other points and, again, are sufficient to predict the presence or absence of the box. 

The CAM method managed to capture both squares most of the times, but by design had limited spatial resolution. Note that due to the lower number of max-pooling layers used for the CAM classifier each pixel in the last feature map had a receptive field of only 39x39 pixels. This could mean that many pixels in that feature map could not simultaneously see both of the squares, which may have contributed to the squares being better discerned. However, we did not investigate this further. 

Lastly, our proposed VA-GAN method produced the most localised disease effect maps, finding the entire boxes and following the edges closely. It also managed to consistently identify both disease effects. 

\begin{table}[h]
 \caption{NCC scores for experiments on synthetic data.}
 \centering
 \begin{tabular}{l l l}
   {\bf Method} & {\bf mean} & {\bf std.} \\
   \midrule
   Guided Backprop \cite{springenberg2014striving}       & 0.14       & 0.04 \\
   Integrated Gradients \cite{sundararajan2017axiomatic} & 0.36       & 0.11 \\
   CAM \cite{zhou2016learning}                           & 0.48       & 0.04 \\
   \midrule
   Additive Perturbation                                 & 0.06       & 0.03 \\
   VA-GAN                                                & {\bf 0.94} & 0.07 \\
 \end{tabular}
 \label{tab:synth_results}
\end{table}

The quantitative NCC results shown in Table \ref{tab:synth_results} are mostly consistent with our qualitative observations, with VA-GAN obtaining significantly higher NCC than the other methods. The additive perturbation technique achieved a low score due to its exclusive focus on edges. 
    
\subsection{Experiments on real neuroimaging data}\label{sec:exp_real}

In this section, we investigate the methods' ability to detect the areas of the brain which are involved in the progression from MCI to AD at a subject-specific level. We trained on images from both categories and then generated disease effect maps only for the AD images. \\ \vspace{-3mm}

\noindent{\bf Data: } We selected 5778 3D T1-weighted MR images from 1288 subjects with either an MCI (label 0) or AD (label 1) diagnosis from the ADNI cohort. 2839 of the images were acquired using a 1.5T magnet, the remainder using a 3T magnet. The subjects are scanned at regular intervals as part of the ADNI study and a number of subjects converted from MCI to AD over the years. We did not use these correspondences for training, however, we took advantage of it for evaluation as will be described later. An overview of the data is given in the supplemental materials in Section~C. 

All images were processed using standard operations available in the FSL toolbox \cite{smith2004advances} in order to reorient and rigidly register the images to MNI space, crop them and correct for field inhomogeneities. We then skull-stripped the images using the ROBEX algorithm \cite{iglesias2011robust}. Lastly, we resampled all images to a resolution of 1.3\,\si{mm^3} and normalised them to a range from -1 to 1. The final volumes had a size of 128x160x112 voxels. \\ \vspace{-3mm}


\noindent{\bf Evaluation: } We split the data on a subject level into a training, testing and validation set containing 825, 256 and 207 subjects, respectively. We then trained all of the algorithms with both AD and MCI data as described earlier, and generated disease effect maps for the AD subjects from the test set. The validation set was used to monitor the training. 

In order to better understand the quality of the generated disease maps we estimated the actual deformations for a number of subjects as follows. We identified all subjects from the test set who were diagnosed with MCI during the baseline examination but progressed to AD in one of the follow-up scans. We then aligned those images rigidly and subtracted them from each other to obtain an \emph{observed disease effect map}. We excluded all subjects which were not acquired with the same field strength, since a large amount of the observed effects could be due to differences in image quality. This left 50 subjects which we evaluated more closely. We note that even for the same field strength there are a number of artefacts due to intensity variations and registration. Furthermore, there are likely to be effects not caused by the disease, such as ageing (which will also be captured by our method), such that the observed disease effect maps could be considered a ground-truth. 

Nevertheless, we also evaluated NCC between the observed and the predicted disease effect maps in the same manner as for the synthetic data. \\ \vspace{-3mm} 

\begin{figure*}[t]
  \centering
    \includegraphics[width=0.94\textwidth]{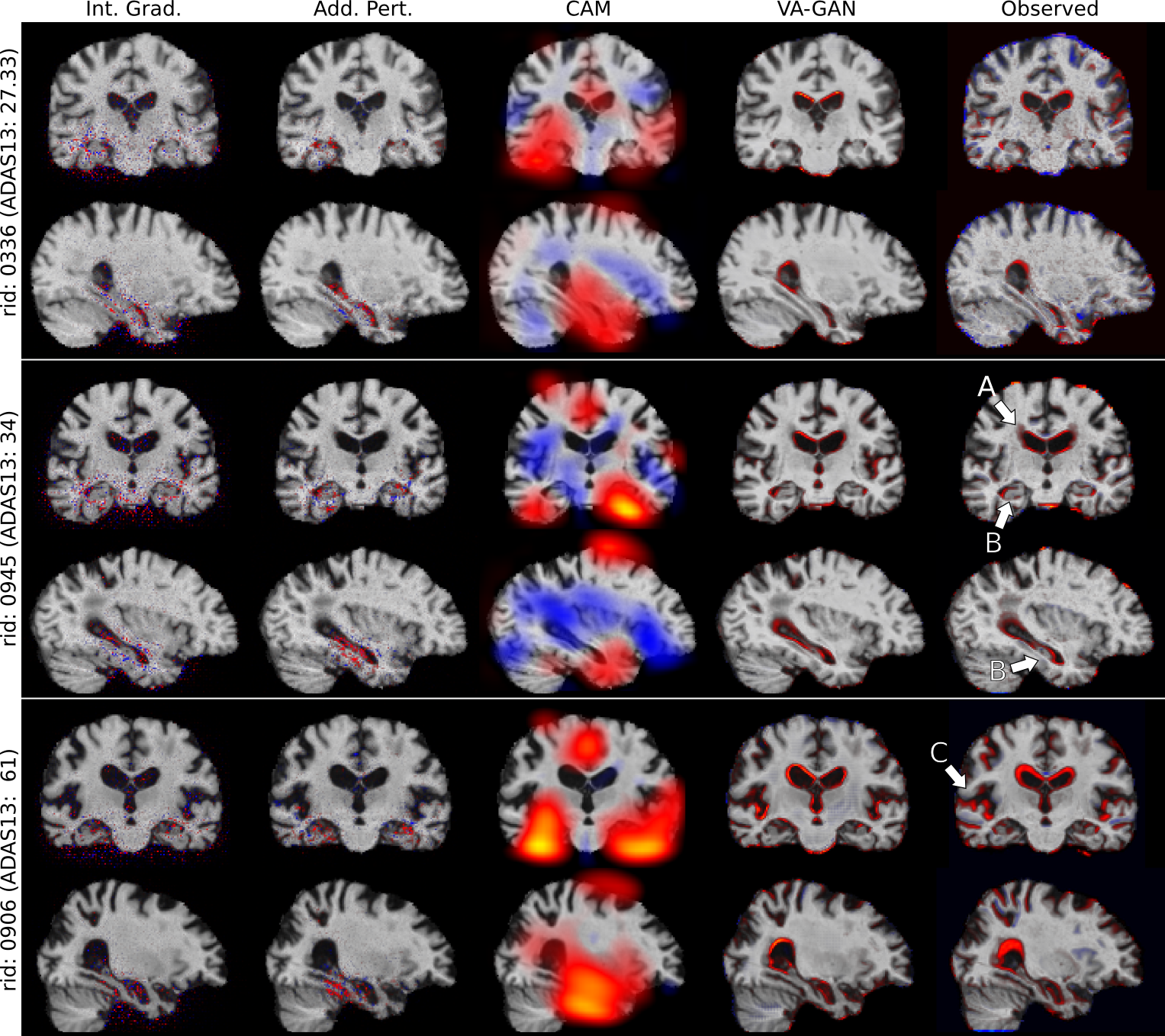} 
    \caption{Coronal and sagittal views of generated AD effect maps for three subjects and actual observed effects. Maps are shown as coloured overlay over the input image. The ventricular (arrow A) and hippocampal (arrow B) regions are particularly affected by the disease and are reliably captured by VA-GAN. In later stages also other brain regions such as the temporal lobe (arrow C) are affected. We also report the ADAS13 cognitive exam scores (larger means AD is further progressed) and the ADNI identifier (rid) for each subject.}
    \label{fig:real_results}
\end{figure*}

\noindent{\bf Results: } Fig. \ref{fig:real_results} shows disease effect maps obtained for a selection of AD subjects (we again omitted Guided Backprop in the figure). The subjects are ordered by increasing progression of the disease as measured by the ADAS13 cognition exam \cite{rosen1984new}. It can be seen that VA-GAN's predictions were in very good agreement with the observed effect maps. As is known from the literature \cite{braak1991neuropathological,dickerson2009cortical} the method indicates atrophy in the hippocampi, and general brain atrophy around the ventricles. Furthermore, it is known that in later stages of the disease other brain areas such as the temporal lobe get affected as well \cite{villemagne2015tau}. Those effects were also identified by VA-GAN in the last subject in Fig. \ref{fig:real_results}. 

\begin{figure}[t]
  \centering
    \includegraphics[width=\columnwidth]{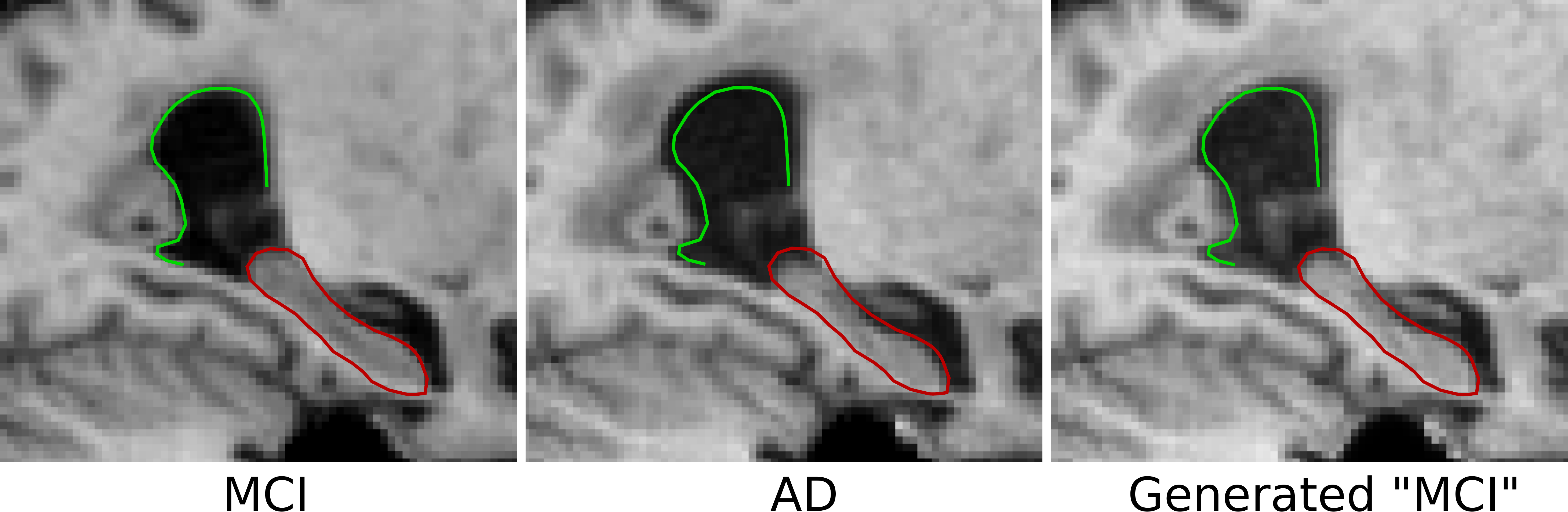}
    \caption{Close-up of the hippocampus region of a subject before (left) and after developing AD (middle). The right panel shows the generated image. The red (hippocampus) and green (ventricles) contours are in the same location in all three images. It can be observed that the map ``reverses'' some of the atrophy.}
    \label{fig:closeup}
\end{figure}

The backprop-based methods and additive perturbations were observed to be very noisy and tended to identify \emph{only} the hippocampal areas. We believe that this is in agreement with the findings on the synthetic data. The hippocampus is known to be the most predictive region for AD, however, it is also known that many other regions are involved in the disease. It is likely, that classifiers learned to focus only on the most discriminative set of features ignoring the rest. Lastly, it is hard to interpret the results produced by CAM due to the low resolution. However, the images suggest that this method focuses on similar areas as the other methods.

Quantitative results are given in Table \ref{tab:real_results}. VA-GAN obtained the highest correlation scores, however, it is hard to draw conclusions from these figures due to the noisy nature of the observed effect maps as well as the possible non-disease related effects on the observed effect maps, which are taken to be ``ground-truth'' in the experiments.

\begin{table}[t]
   \caption{NCC scores for experiments on neuroimaging data.}
   \centering
   \begin{tabular}{l l l}
     {\bf Method} & {\bf mean} & {\bf std.} \\
     \midrule
     Guided Backprop \cite{springenberg2014striving}       & 0.05       & 0.03 \\
     CAM \cite{zhou2016learning}                           & 0.09       & 0.07 \\
     Integrated Gradients \cite{sundararajan2017axiomatic} & 0.13       & 0.05 \\
     \midrule
     Additive Perturbation                                 & 0.11       & 0.05 \\
     VA-GAN                                                & {\bf 0.27} & 0.15 \\
   \end{tabular}
   \label{tab:real_results}
\end{table}

We observed that VA-GAN generally produced very realistic deformations. In Fig. \ref{fig:closeup} a close-up of the MCI, AD, and generated image is shown for a sample subject. It can be seen that our method succeeded in making the generated image more similar to the corresponding MCI image and that the changes were realistic.

\section{Limitations and discussion}

We have proposed a method for visual feature attribution using Wasserstein GANs. It was shown that, in contrast to backprop-based methods, our technique can capture multiple regions affected by disease, and produces state-of-the-art results for the prediction of disease effect maps in neuroimaging data and on a synthetic dataset. 

Currently, the method assumes that the category labels of the test data are known during test-time. In case they are unknown, the method could be easily combined with classifier which produces this information. We only evaluated the method for the case of two labels. More categories could be addressed by training multiple map generators each mapping to a background class (assuming there is one). 

In the future, we plan to model other effects such as ageing or the presence or absence of certain genes on the ADNI data, investigate the method on other datasets and apply it to other problems such as weakly-supervised localisation. 

\subsection*{Acknowledgements}
We gratefully acknowledge the support of NVIDIA Corporation with the donation of a Titan Xp GPU. 

{\small
\bibliographystyle{ieee}
\bibliography{references}
}

\newpage
\begin{appendices}

\section{Network architectures}

In this section we describe the exact network architectures used for the 3D VA-GAN. We present the critic and map generator functions as Python-inspired pseudo code, which we found easier to interpret than a graphical representation. The layer parameters are specified as arguments to the layer functions. Unless otherwise specified all convolutional layers used a stride of 1x1x1 and a rectified linear unit (ReLU) non-linearity. 

The architecture of the critic function $D(x)$ is shown in Fig. \ref{fig:critic}. The \texttt{conv3D\char`_layer} function performs a regular 3D convolution without batch normalisation and the \texttt{global\char`_averagepool3D} function performs an averaging over the spatial dimensions of the feature maps. 

The architecture for the map generator function $M(x)$ is shown in Fig. \ref{fig:generator}. Here, the \texttt{conv3D\char`_layer\char`_bn} is a 3D convolutional layer with batch normalisation before the nonlinearity. The \texttt{deconv3D\char`_layer\char`_bn} learns an upsampling operation as in the original U-Net and also uses batch normalisation. Lastly, the \texttt{crop\char`_and\char`_concat\char`_layer} implements the skip connections across the bottleneck by stacking the feature maps along the dimension of the channels. 

Note that the architectures for the 2D experiments on synthetic data were identical, except all 3D operations were replaced by their 2D equivalents. 

\begin{figure*}
\begin{lstlisting}[language=Python]

def critic(x):

    # inputs
    #    x:      an image from category c=0, or an image from category c=1
    #            plus the additive mask M(x)
    # returns
    #    logits: the critic output for x 

    conv1_1 = conv3D_layer(x, num_filters=16, kernel_size=(3,3,3))

    pool1 = maxpool3D_layer(conv1_1)

    conv2_1 = conv3D_layer(pool1, num_filters=32, kernel_size=(3,3,3))

    pool2 = maxpool3D_layer(conv2_1)

    conv3_1 = conv3D_layer(pool2, num_filters=64, kernel_size=(3,3,3))
    conv3_2 = conv3D_layer(conv3_1, num_filters=64, kernel_size=(3,3,3))

    pool3 = maxpool3D_layer(conv3_2)

    conv4_1 = conv3D_layer(pool3, num_filters=128, kernel_size=(3,3,3))
    conv4_2 = conv3D_layer(conv4_1, num_filters=128, kernel_size=(3,3,3))

    pool4 = maxpool3D_layer(conv4_2)

    conv5_1 = conv3D_layer(pool4, num_filters=256, kernel_size=(3,3,3))
    conv5_2 = conv3D_layer(conv5_1, num_filters=256, kernel_size=(3,3,3))
    conv5_3 = conv3D_layer(conv5_2, num_filters=256, kernel_size=(3,3,3))

    conv5_4 = conv3D_layer(conv5_3, 
                           num_filters=1, 
                           kernel_size=(1,1,1),
                           nonlinearity=identity)

    logits = global_averagepool3D(conv5_4)

    return logits

\end{lstlisting}
\caption{VA-GAN critic architecture. }
\label{fig:critic}
\end{figure*}

\begin{figure*}
\begin{lstlisting}[language=Python]

def map_generator(x):

    # inputs
    #    x: an image from category c=1
    # returns
    #    M: additive map M(x) such that y = x + M(x) appears to be from c=0

    # Encoder: 

    conv1_1 = conv3D_layer_bn(x, num_filters=16, kernel_size=(3,3,3))
    conv1_2 = conv3D_layer_bn(conv1_1, num_filters=16, kernel_size=(3,3,3))
    
    pool1 = maxpool3D_layer(conv1_2)

    conv2_1 = conv3D_layer_bn(pool1, num_filters=32, kernel_size=(3,3,3))
    conv2_2 = conv3D_layer_bn(conv2_1, num_filters=32, kernel_size=(3,3,3))

    pool2 = maxpool3D_layer(conv2_2)

    conv3_1 = conv3D_layer_bn(pool2, num_filters=64, kernel_size=(3,3,3))
    conv3_2 = conv3D_layer_bn(conv3_1 num_filters=64, kernel_size=(3,3,3))

    pool3 = maxpool3D_layer(conv3_2)

    # Bottleneck:

    conv4_1 = conv3D_layer_bn(pool3, num_filters=n128, kernel_size=(3,3,3))
    conv4_2 = conv3D_layer_bn(conv4_1, num_filters=128, kernel_size=(3,3,3))

    # Decoder: 

    upconv3 = deconv3D_layer_bn(conv4_2, kernel_size=(4,4,4), strides=(2,2,2), num_filters=64)
    concat3 = crop_and_concat_layer([upconv3, conv3_2])

    conv5_1 = conv3D_layer_bn(concat3, num_filters=64, kernel_size=(3,3,3))
    conv5_2 = conv3D_layer_bn(conv5_1, num_filters=64, kernel_size=(3,3,3))

    upconv2 = deconv3D_layer_bn(conv5_2, kernel_size=(4,4,4), strides=(2,2,2), num_filters=32)
    concat2 = crop_and_concat_layer([upconv2, conv2_2])

    conv6_1 = conv3D_layer_bn(concat2, num_filters=32, kernel_size=(3,3,3))
    conv6_2 = conv3D_layer_bn(conv6_1, num_filters=32, kernel_size=(3,3,3))

    upconv1 = deconv3D_layer_bn(conv6_2, kernel_size=(4,4,4), strides=(2,2,2), num_filters=16)
    concat1 = crop_and_concat_layer([upconv1, conv1_2])

    conv8_1 = conv3D_layer_bn(concat1, num_filters=16, kernel_size=(3,3,3))

    M = conv3D_layer(conv8_1, 
                     num_filters=1, 
                     kernel_size=(3,3,3), 
                     nonlinearity=identity)

    return M

\end{lstlisting}
\caption{VA-GAN map generator architecture.}
\label{fig:generator}
\end{figure*}

\section{Close-up analysis of VA-GAN}

\begin{figure*}[t]
  \centering
    \includegraphics[width=0.96\textwidth]{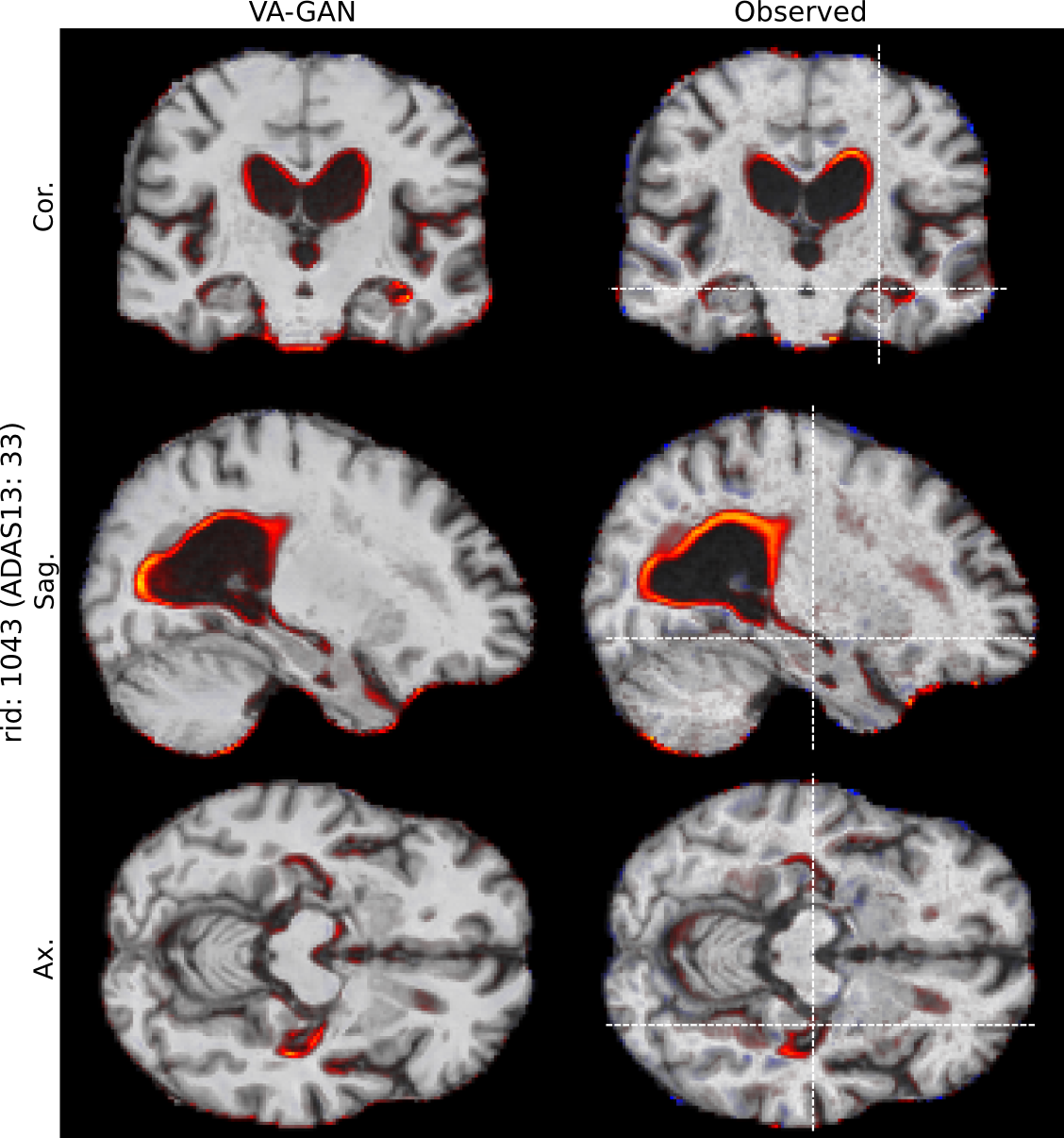} 
    \caption{Coronal, sagittal and axial views of the predicted and observed disease effect maps for an additional subject. The location of the planes is indicated by dotted white lines in the right column. In order to allow for an enlarged view, only the predictions obtained by VA-GAN are shown. The ADNI rid and the ADAS13 score for this subject are reported on the left-hand side.}
    \label{fig:three_views}
\end{figure*}

In Fig. \ref{fig:three_views} we present a larger view of all three orthogonal planes for an additional subject. In order to allow for an enlarged view, we only include the results obtained by VA-GAN and the actual observed changes from MCI to AD. As before it can be seen that VA-GAN produced visual attribution maps that very closely approximate the observed deformations. In particular, we note that for this subject VA-GAN correctly predicted a smaller disease effect in the left hippocampus compared to the right hippocampus. 

\section{Details of MR brain data cohort}

The MR brain image data used in preparation of this article were obtained from the Alzheimer’s Disease
Neuroimaging Initiative (ADNI) database (adni.loni.usc.edu). As such, the investigators
within the ADNI contributed to the design and implementation of ADNI and/or provided data
but did not participate in analysis or writing of this report. A complete listing of ADNI
investigators can be found at:
\url{http://adni.loni.usc.edu/wp-content/uploads/how_to_apply/ADNI_Acknowledgement_List.pdf}.

\begin{figure}
    \centering
    \includegraphics[width=1.00\columnwidth]{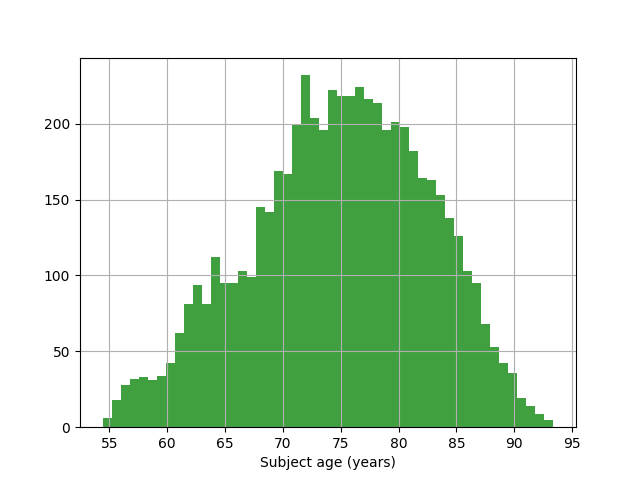} 
    \caption{Histogram of the subject age of all ADNI images used in this work. The mean age was $74.89$ years, with a standard deviation of $7.70$.}
    \label{fig:age_hist}
\end{figure}

Specifically, we used T1-weighted MR data from the ADNI1, ADNIGO and ADNI2 cohorts which were acquired in with a mixture of 1.5T and 3T scanners. The data consisted of 5770 images, acquired from 1291 subjects. The images for each subject were acquired at separate visits that were spaced in regular intervals from 6 months to one year and usually spanned multiple years. On average each subject was scanned 4.5 times. The cohort consisted of 496 female and 795 male subjects. 2839 of the images were acquired using a 1.5T magnet, the remainder using a 3T magnet. The distribution of the ages at which the images were acquired is shown in Fig. \ref{fig:age_hist}. We only considered images with a diagnosis of mild cognitive impairment (MCI) or Alzheimer's disease (AD).

After preprocessing we randomly divided the data into a training, testing and validation set. We performed the split on a subject basis rather than an image basis. The exact split is shown in Table \ref{tab:datasplit}. The table furthermore shows the distribution over the diagnoses on a image level, and the number of subjects which have undergone a conversion from MCI to AD in the examined time intervals. 

The training data was used for learning the mask generator and critic parameters which minimise the cost function in Eq. 4 of the main article. The validation set was used for monitoring of the training based on the Wasserstein distance and visual examination of generated masks, and for hyperparameter tuning. The test set was used for the final qualitative and quantitative evaluation. 

\begin{table}[t]
   \caption{Detailed information on data split into training, testing and validation data.}
   \centering
    \begin{tabular}{l | c c c | c}
    \hline
    ~ & Train & Test & Validation & Total \\
    \hline
    Num. Imag. & ~ & ~ & ~ & ~ \\
    ~~ MCI     & 2520 & 755  & 639 & 3914 \\
    ~~ AD      & 1199 & 399  & 266 & 1864 \\
    ~~ Total   & 3719 & 1154 & 905 & 5778 \\
    \hline
    Num. Subj.        & ~   & ~   & ~   & ~ \\
    ~~ Converters     & 172 & 51  & 49  & 272 \\
    ~~ Non-converters & 653 & 208 & 158 & 1019 \\
    ~~ Total          & 825 & 259 & 207 & 1291\\
    \hline
    \end{tabular}
    \label{tab:datasplit}
\end{table}

In case of interest, a list of the exact ADNI subject ID's used in the study can be found in our public code repository (\url{https://github.com/baumgach/vagan-code}) in the folder {\tt data/subject\_rids.txt}.

\section{Alternative classifier architecture}

It was suggested during the reviews that our classifier architecture with two dense layers before the final output is responsible for the poor performance of the backpropagation based saliency map techniques. It was recommended that we investigate the popular class of architectures where the final convolutions are aggregated using a global average pooling step over the spatial dimensions of the activation maps, followed by a single dense layer. Examples of this type of architecture include the works of He at al. \cite{he2016deep} and Lin et al. \cite{lin2013network}. In our experiments, the class activation mappings (CAM) method \cite{zhou2016learning} was also using this general architecture. In theory this may abstract the data less before the final output and perhaps produce maps that can more easily identify multiple regions in the image. 

\begin{figure}
    \centering
    \includegraphics[width=1.00\columnwidth]{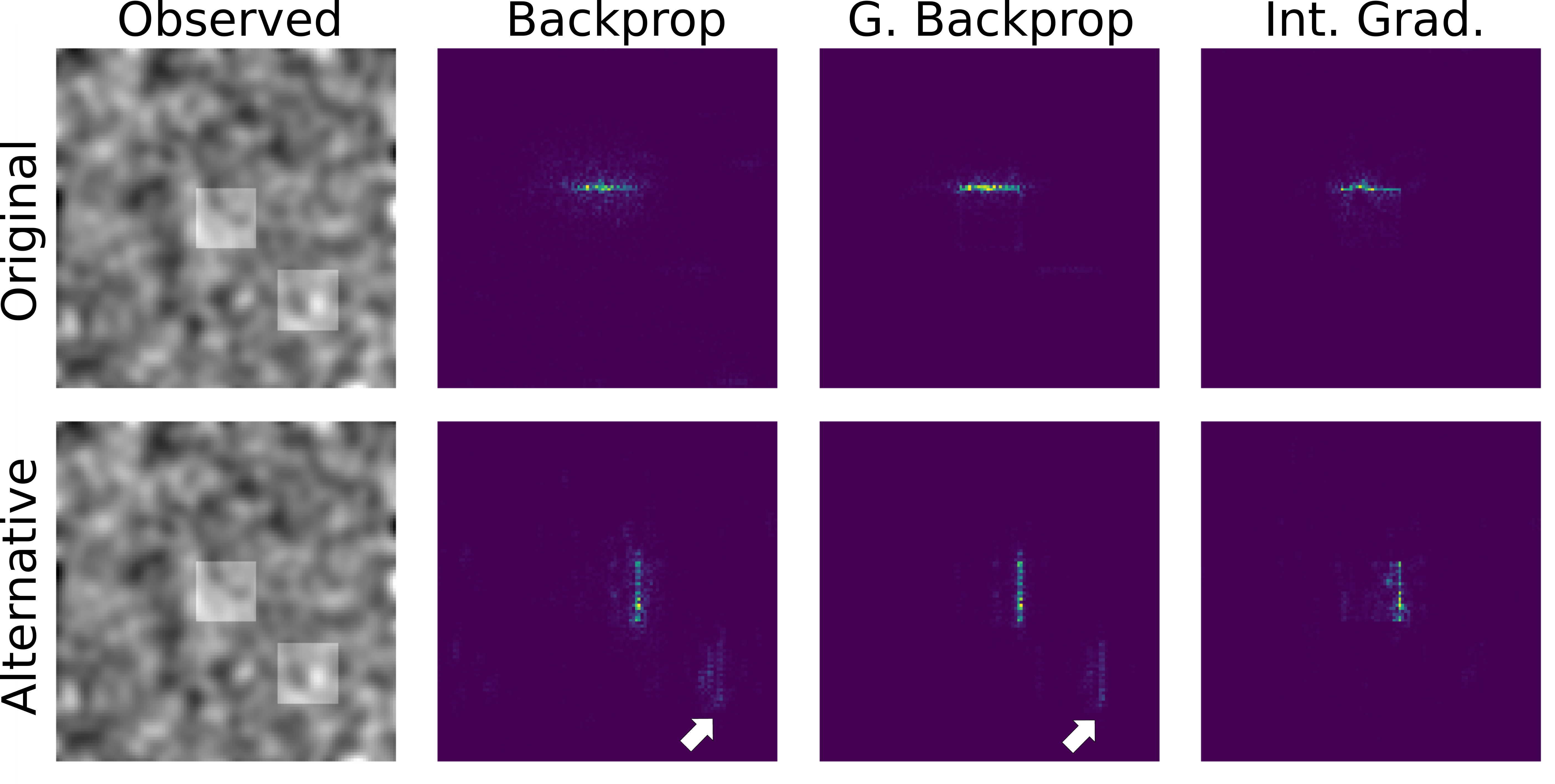} 
    \caption{Saliency maps obtained using simple backpropagation, guided backpropagation and integrated gradients for two different network architectures: (1) the original architecture from the synthetic experiments (Section 4.2) in the main article, (2) an alternative architecture with a global average pooling layer followed by a single dense layer before the final classification output. The white arrows in the second row highlight very faint attributions of the second box.}
    \label{fig:rebuttal_saliencies}
\end{figure} 

To investigate this theory we repeated the synthetic experiment (outlined in Section 4.2 of the main article), but replaced the final two dense layers in our synthetic experiments by a global average pooling and a single dense layer. After full convergence of the network from the main article and the alternative architecture, we obtained the saliency maps shown in Fig. \ref{fig:rebuttal_saliencies}. In addition to the integrated gradients method \cite{sundararajan2017axiomatic} already shown in the main article, here we also show the results for normal backprop \cite{simonyan2013deep} and guided backprop \cite{springenberg2014striving}. It can be observed that indeed, with the alternative architecture, normal and guided backprop manage to correctly attribute some of the pixels of the peripheral box, albeit very faintly (emphasised with white arrows in Fig \ref{fig:rebuttal_saliencies}). However, regardless of the architecture the classifier appears to focus only on the pixels of one of the edges, which is only subset of the features characterising this class. Note that the orientation of the attributed edges depends on the random initialisation of the network.

\begin{figure}
    \centering
    \includegraphics[width=1.00\columnwidth]{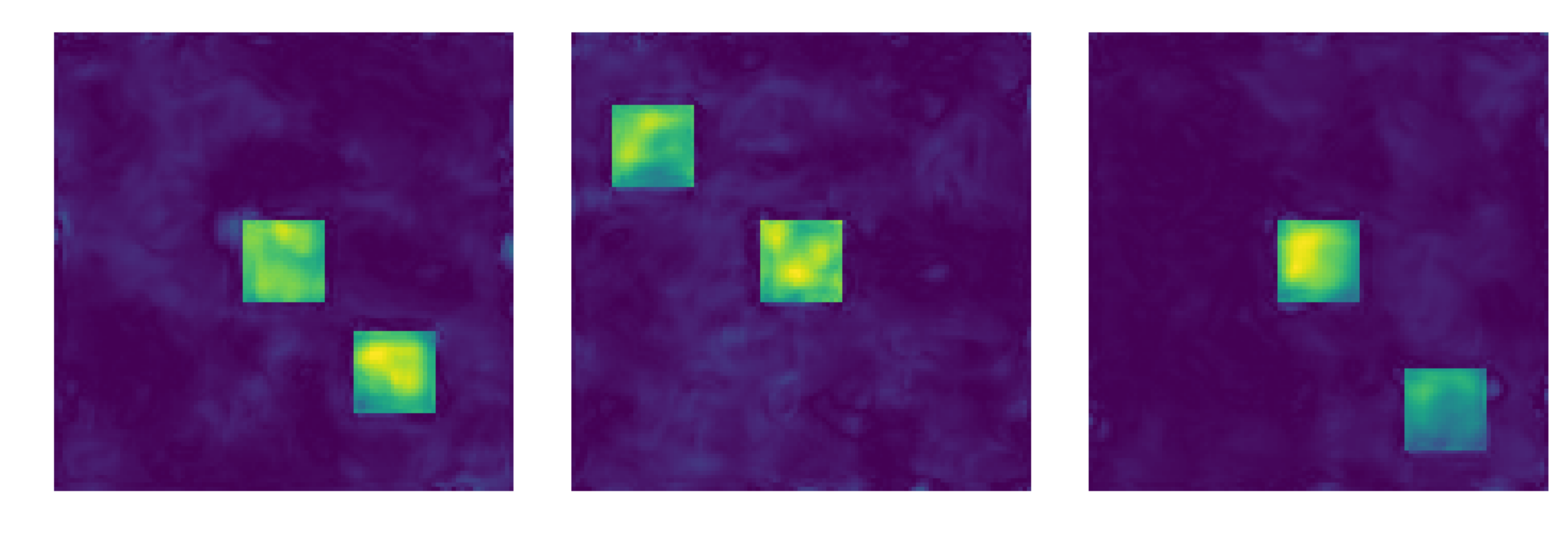} 
    \caption{Visual feature attribution maps obtained using our proposed VA-GAN method. The first sample corresponds to the input image in Fig. \ref{fig:rebuttal_saliencies}. The other two images correspond to other random input images.}
    \label{fig:vagan_samples}
\end{figure}

Nevertheless, the feature attribution maps obtained using the backprop-based techniques are not of comparable quality to the maps produced by our proposed VA-GAN method. For emphasis we show the corresponding feature attribution map produced with VA-GAN plus two more samples in Fig.~\ref{fig:vagan_samples}.

To conclude, we would like to note that from the point of view of saliency maps, (1) two dense layers or (1) average pooling followed by a dense layer, are conceptually similar. In both cases the final prediction aggregates information from multiple receptive fields covering the whole image. Therefore, it is not surprising that the two networks behave similarly. As outlined in the work of Shwartz-Ziv et al. \cite{shwartz2017opening} the optimisation of neural network classifiers results in a trade off between compression of input features and predictive accuracy. In both networks, the final prediction has access to all features in the image and thus has the potential to compress away features that are redundant for classification (such as one of the two boxes). 

\end{appendices}

\end{document}